\documentclass[letterpaper, 10 pt, conference]{ieeeconf-modified}
\IEEEoverridecommandlockouts                              

\usepackage{graphics}
\usepackage{amsmath} 
\usepackage{amssymb}

\usepackage{mathtools}

\usepackage[numbers]{natbib}
\usepackage{booktabs}
\usepackage{hyperref}
\usepackage[capitalize]{cleveref}

\usepackage{siunitx}

\crefname{equation}{}{}  

\newcommand{\gitlink}{https://dlr-alr.github.io/2023-humanoids-ik/}
\DeclareMathOperator*{\argmin}{argmin}

\newcommand{\NDoF}{N_{\mathrm{DoF}}}
\newcommand{\Nbasisset}{N_{\mathrm{b}}} 
\newcommand{\Nframes}{N_{\mathrm{f}}} 
 
\newcommand{\Nspheresi}{N_{\mathrm{s}i}} 
\newcommand{\Nspheresj}{N_{\mathrm{s}j}} 

\newcommand{\SOThree}{SO(3)}
\newcommand{\RThree}{\mathbb{R}^3}
\newcommand{\RDoF}{\mathbb{R}^{\NDoF}}

\newcommand{\q}{q}
\newcommand{\qdefault}{\Bar{q}}

\newcommand{\Frames}{F}
\newcommand{\FrameTrue}{\Bar{F}}
\newcommand{\pos}{p}
\newcommand{\Rot}{R}

\newcommand{\funkin}{f}
\newcommand{\funclip}{c}
\newcommand{\distfield}{D}

\newcommand{\CostTotal}{U}
\newcommand{\CostRest}{U_\mathrm{A}}

\newcommand{\CostLength}{U_\mathrm{L}}
\newcommand{\CostWorldCollision}{U_\mathrm{W}}
\newcommand{\CostSelfCollision}{U_\mathrm{S}}
\newcommand{\CostFrame}{U_\mathrm{F}}
\newcommand{\CostFramePos}{U_\mathrm{P}}
\newcommand{\CostFrameRot}{U_\mathrm{\Rot}}

\newcommand{\lambdaLength}{\lambda_\mathrm{L}}
\newcommand{\lambdaWorldCollision}{\lambda_\mathrm{W}}
\newcommand{\lambdaSelfCollision}{\lambda_\mathrm{S}}

\newcommand{\lambdaFrameRot}{\lambda_\mathrm{\Rot}}

\newcommand{\Spheres}{\boldsymbol{S}} 

\newcommand{\xsphere}{x} 
\newcommand{\rsphere}{r}

\newcommand{\BPS}{B}
\newcommand{\bps}{b}

\newcommand{\NetWeights}{\Theta}

\newcommand{\xWorld}{x_{\mathrm{W}}} 
\newcommand{\xFrame}{x_{\mathrm{F}}}

\newcommand{\StaticArm}{Flat Arm}
\newcommand{\JustinArm}{LWR III}
\newcommand{\Justin}{Agile Justin}

\usepackage{fancyhdr}
\fancypagestyle{FirstPage}{
\lfoot{\fontsize{7}{7}\selectfont \begin{center}\copyright2023 IEEE. Personal
use of this material is permitted. \\Permission from IEEE must be
obtained for all other uses, in any current or future media,
including reprinting/republishing this material for advertising or
promotional purposes, creating new collective works, for resale or
redistribution to servers or lists, or reuse of any copyrighted
component of this work in other works.\end{center}}
}

\title{\LARGE \bf
Efficient Learning of Fast Inverse Kinematics with Collision Avoidance
}
\author{Johannes Tenhumberg$^{*1,2,3}$ \;\; Arman Mielke$^{*1,3}$ \;\; Berthold Bäuml$^{1,2}$
\thanks{$^{1}$DLR Institute of Robotics \& Mechatronics, Germany;
$^{2}$Deggendorf Institute of Technology, Germany; 
$^{3}$Technical University of Munich, Germany}
\thanks{$^{*}$First two authors contributed equally.}
\thanks{This work was partly funded by the Bavarian Ministry of Economic Affairs, Regional Development and Energy, within the projects SMiLE2gether (LABAY102).}
\thanks{Web: (\href{\gitlink}{\gitlink})}
\thanks{Contact: \tt\footnotesize johannes.tenhumberg@dlr.de}
}

\begin{document}

\maketitle
\thispagestyle{empty}
\pagestyle{empty}


\begin{abstract}
Fast inverse kinematics (IK) is a central component in robotic motion planning.
For complex robots, IK methods are often based on root search and non-linear optimization algorithms.
These algorithms can be massively sped up using a neural network to predict a good initial guess, which can then be refined in a few numerical iterations.
Besides previous work on learning-based IK, we present a learning approach for the fundamentally more complex problem of IK with collision avoidance. 
We do this in diverse and previously unseen environments.
From a detailed analysis of the IK learning problem, we derive a network and unsupervised learning architecture that removes the need for a sample data generation step.
Using the trained network's prediction as an initial guess for a two-stage Jacobian-based solver allows for fast and accurate computation of the collision-free IK. 
For the humanoid robot, Agile Justin (19 DoF), the collision-free IK is solved in less than \SI{10}{\milli\second} (on a single CPU core) and with an accuracy of \SI{e-4}{\meter} and \SI{e-3}{\radian} based on a high-resolution world model generated from the robot's integrated 3D sensor.
Our method massively outperforms a random multi-start baseline in a benchmark with the 19 DoF humanoid and challenging 3D environments. 
It requires ten times less training time than a supervised training method while achieving comparable results.
\end{abstract}

\section{Introduction}
\thispagestyle{FirstPage}
A solution to inverse kinematics (IK) while avoiding collisions is fundamental for getting joint configurations in the most common robotic tasks, such as picking and placing objects. 
Still, it can also be used in the context of motion planning:
For example, positioning a cup upright on a cluttered table requires motions with cartesian constraints at the end-effector.
This joint problem of solving the IK and getting a collision-free trajectory is challenging, so it is often divided into two sub-problems~\cite{Schulman2014}.
First, the IK is solved to find the final configuration for grasping the object, and then the trajectory from the initial configuration to the goal configuration is planned.
Similarly, some problems require a path in which the frame of the end-effector is constrained at some intermediate steps.
Solving the IK problems along the path and initializing the motion planner with the solutions can benefit these problems.
Therefore, quickly computing a collision-free IK is crucial for real-time grasping and manipulation tasks.

\begin{figure}[t]
    \centering
	\includegraphics[width=0.9\linewidth]{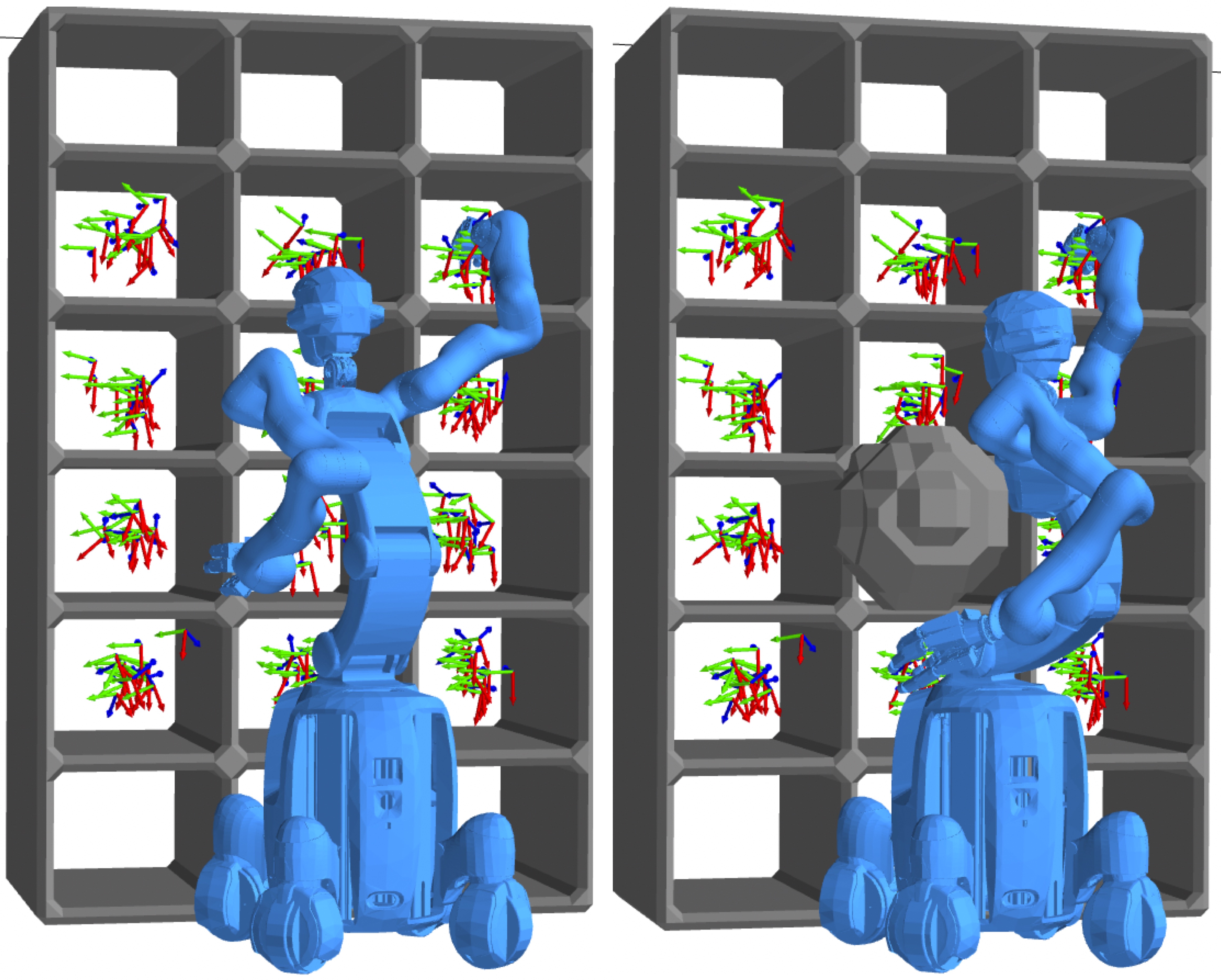}
	\caption{DLR's Agile Justin~\cite{Bauml2014} in a shelf environment. 
    The frames for the IK problem were sampled randomly in the respective boxes of the shelf (left). 
    On the right, the previous solution was blocked and made infeasible by placing an additional obstacle in the workspace, leading to a different collision-free solution.
    Details about the training datasets, networks, and videos can be found on the paper's website.}
    \label{fig:Justin19_shelf}
    \vspace{-0.5cm}
\end{figure}

In this paper, we deal with the problem of speeding up the IK with collision avoidance via learning for complex robots like  DLR's humanoid robot Agile Justin~\cite{Bauml2014} with 19 degrees of freedom (DoF) as depicted in \cref{fig:Justin19_shelf}. 
As we will show, learning the inherently ambiguous inverse mapping from a frame of a robot's TCP (tool center point) to its joint configuration poses several challenges (other than, e.g., in speeding up motion planning in configuration space~\cite{Tenhumberg2022}). 
Learning an IK solution gets especially hard when incorporating self-collision avoidance and avoiding collisions with obstacles in arbitrary environments (see \cref{fig:Justin19_shelf}, right). 

This paper presents and compares two learning-based approaches to the IK problem: a supervised learning approach that relies on a separate data generation step with boosting and an unsupervised approach that does not need time-consuming data generation and works directly on the objective function.

\subsection{Related Work}

There are many non-learning-based methods to solve the IK problem.
They are often based on the Inverse Jacobian method~\cite{Sukavanam2011, Sugihara2011, Colome2015}.
A popular algorithm is TRAC-IK~\cite{Beeson2015}, combining a Newton-based algorithm with a Sequential Quadratic Programming (SQP) method.
Running both methods in parallel and terminating if one succeeds improves speed and robustness.

Other optimization methods have been applied to the IK problem, too.
One is Particle Swarm Optimization (PSO), which \citet{Collnism2017PSO} used for snake-like robots with many degrees of freedom (DoF) and \citet{Rokbani2013PSO} analyzed for a two-link robot statistically.
\citet{Trutman2022} describe the IK as a polynomial optimization problem, which they use to find a globally optimal solution for serial 7\,DoF robots.
\citet{Tringali2020} leverage the Inverse Jacobian method but use a randomized matrix to weight the pseudo inverse.
This adaptation allows them to improve the convergence to a globally optimal IK solution.
However, none of these methods take into consideration collisions with the environment.

\citet{Ferrentino2021} uses dynamic programming to solve the IK with obstacles and make it available in ROS~\cite{ROS2022}.
\citet{Giamou2022} formulate the IK problem as a distance-geometric problem, allowing them to use semidefinite programming methods to find a low-rank solution.
While the approach is elegant and fast, it can only incorporate spherical obstacles.
\citet{Zhao2021} introduce a modern and fast solver that combines Inverse Jacobian methods, SQP, and PSO.
Their method can handle dynamic obstacles but is also limited to spheres.

There are also many learning-based approaches for solving the IK problem more efficiently.
One problem for all those methods is the inherent ambiguity of the IK solution.
\citet{Bocsi2011} tackles ambiguous solutions by using structured prediction.
\citet{Kim2021} uses the graph structure of robot kinematics to learn the entire nullspace with a Graph Neural Network (GNN).
However, the nullspace gets exponentially large with the DoFs of the robot, making it crucial that the learning of the mapping between forward and inverse kinematics is efficient~\cite{Kubus2018, Ren2020}.
To improve the speed and portability of IK methods, \citet{Zaidel2021} introduce a neuromorphic approach that they apply to a 7\,DoF robot arm.

All IK methods described so far consider an obstacle-free working environment. \citet{Lehner2022} leverage transfer learning between similar robot kinematics in a single environment with obstacles.
They use the network predictions to guide a Rapid Random Tree (RRT) motion planner.
\citet{Lembono2021} use Generative Adversarial Networks (GANs) to learn constrained robot configurations.
They use the predictions of those networks as an initial guess for an optimization-based planner to warm-start the IK problem and as samples for an RRT motion planner.
They consider the environment for their tasks, but for each new scene, an ensemble of GANs needs to be trained to counteract the mode collapse and produce valuable samples.

Until now, no learning-based approach to the IK problem incorporates collision avoidance for arbitrary, previously unseen environments.
Moreover, for autonomous robots, the environment model is generated in real-time from sensor data and, hence, is a high-resolution, e.g., voxel-based model~\cite{Wagner2013}. 
For those challenging worlds, no fast and efficient collision-free IK solver exists.

\subsection{Contributions}

We tackle the problem by formulating the IK with collision avoidance as an optimization problem similar to CHOMP~\cite{Zucker2013} and use a combination of Jacobian-based projections and gradient descent to solve it numerically.
We use the predictions of a neural network as warm-starts to speed up the optimizer.
Those networks are trained with the Basis Point Set (BPS)~\cite{Prokudin2019} as encoding for the environment to incorporate collision avoidance. 
The BPS encoding has already been successfully used for planning robot motions in configuration space~\cite{Tenhumberg2022}.

Our main contributions are:
\begin{itemize}
    \item A learning-based fast and accurate solver for IK with collision avoidance for complex previously unseen environments (for the 19 DoF humanoid Agile Justin on a high-resolution voxel grid an IK solution with an accuracy of \SI{e-4}{\meter} and \SI{e-3}{\radian} in \SI{10}{\milli\second}).
    \item A detailed analysis of the challenges in learning ambiguous IK with collision avoidance and the resulting network, including a twin-headed architecture, a singularity-free output representation, and boosting.
    \item{The optimal solution to the IK problem varies not smoothly across the workspace. 
    We show that two heads are enough for a network to predict the sharp switches between those regions of different modes.}
    \item A benchmark of the supervised and the unsupervised learning approach shows a ten times faster training time for the latter and a more straightforward training procedure while outperforming the random baseline significantly.
\end{itemize}

\section{Optimization-based Inverse Kinematics}\label{sec:IK}

\begin{figure}[t]
    \centering
	\includegraphics[width=\linewidth]{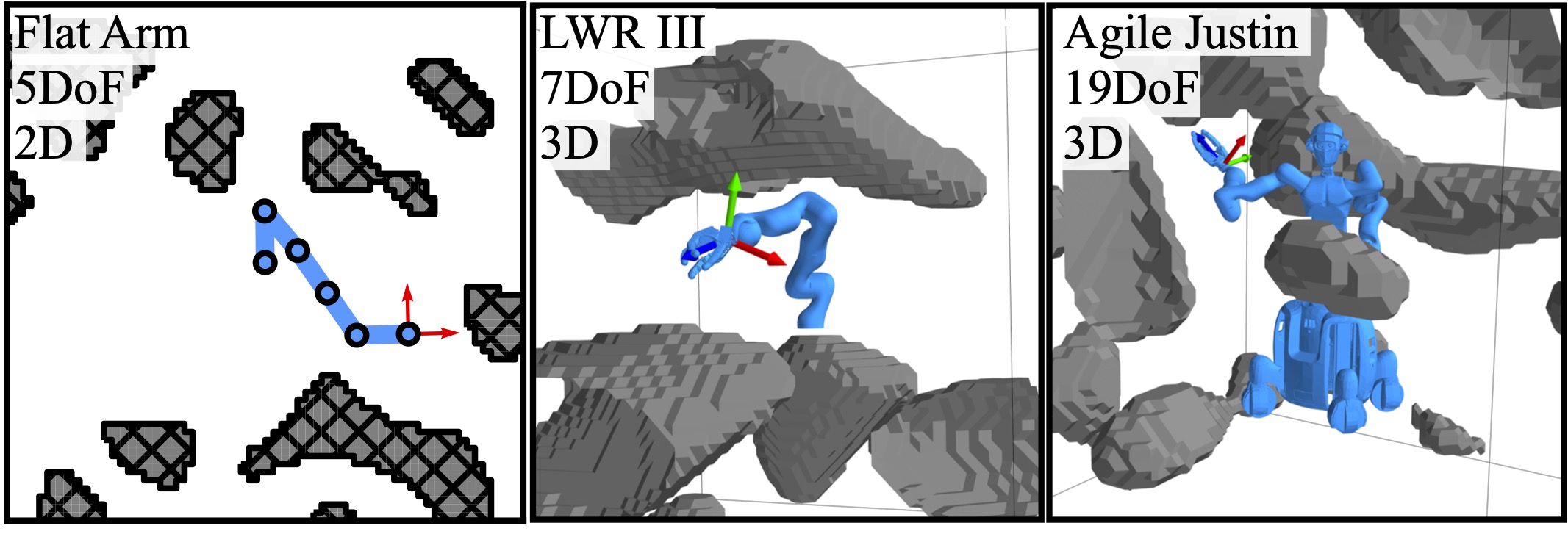}
    \caption{The three robots used in the experiments in environments generated with Simplex Noise~\cite{Perlin2001}.
    The Flat Arm in 2D helps to analyze and visualize the IK problem in detail. The LWR III and Agile Justin demonstrate the capabilities of our method for complex robotic systems.}
	\label{fig:ThreeRobots}
\end{figure}

\subsection{Objective Terms}
We formulate the IK problem, including avoiding collision with the environment and self-collision, as an optimization problem.
These hard constraints are taken into account as weighted terms in an overall objective function.
Central to the problem formulation is the model of the robot.
The forward kinematics maps from joint configurations $\q \in \RDoF$ to the link frames $\{ \Frames_i \}_{i=1}^{\Nframes} = \funkin(\q)$.
Each homogenous transformation matrix $\Frames_i$ describes a full 6D pose $(\pos_i, \Rot_i) \in \RThree \times \SOThree$.

The equality constraint for the IK is that the distance between a specific frame in the chain $\Frames_i$ and a target frame $\FrameTrue_i$ is zero.
In the objective, this results in a translational part
\begin{align}
\CostFramePos(\q, \FrameTrue_i) = \frac{1}{2} \| \pos_i(\q) - \Bar{\pos}_i \|^2
\label{eq:CostFramePos}
\end{align}
 and a  rotational part
\begin{align}
\CostFrameRot(\q, \FrameTrue_i) = \frac{1}{2} ( 3 - \text{Trace}(\, \Rot_i(\q) \, {\Bar{\Rot}_i}^{-1} \,) ).
\label{eq:CostFrameRot}
\end{align}

The goal is to obtain a collision-free IK.
The collision inequality constraint between the robot and the environment is also incorporated in the objective using the following robot and environment models.
We describe the robots geometry of each link $\Frames_i$ by a set of spheres $\Spheres_i=\{\xsphere_{ik}, \rsphere_{ik}\}_{k=1}^{\Nspheresi}$ with centers and radii.
For the world, we use a Signed Distance Field (SDF) $\distfield(x)$, which gives the distance to the closest obstacle for each point $x$ in the workspace.
The collision cost is then given by the sum of all the collisions of the different body parts
\begin{align}
\CostWorldCollision(q) &= \sum_{i=1}^{\Nframes} \sum_{k=1}^{\Nspheresi} 
\funclip\Big( 
    \distfield\big( 
        \Frames_i(\q) \cdot{} \xsphere_{ik}
    \big) 
    - \rsphere_{ik}
\Big).
\label{eq:CostWorldCollision}
\end{align}
In addition to collisions with the world, complex robots must also account for self-collision.
Again, the cost sums up all the penetrations between the different body pairs 
\begin{align}
\CostSelfCollision(q) \!=\! \sum_{j>i}^{\Nframes, \Nframes} \sum_{k, l}^{\Nspheresi, \Nspheresj}
\!\!\!\!\funclip\Big( 
    \big\| 
         \Frames_{i}(\q) \!\cdot\! \xsphere_{ik}  \text{-}  \Frames_{j}(\q) \!\cdot\! \xsphere_{jl} 
    \big\|
    \text{-} \rsphere_{ik} \text{-} \rsphere_{jl}
\Big).
\label{eq:CostSelfCollision}
\end{align}
The smooth clipping function $\funclip$ is introduced to transform the inequality into an equality constraint, which is then written as an additional cost term in the objective~\cite{Schulman2014, Tenhumberg2022}.
It considers only the parts of the robot that are in collision by setting positive distances to zero.
Thus, a collision-free solution has a cost of zero.

While the mapping from the joint configuration to the end-effector frame is unique, the same does not hold for the inverse mapping.
For an over-actuated robot, infinitely many joint configurations can reach a given frame in the workspace.
However, one is usually not interested in an arbitrary solution but one which satisfies additional criteria.
We introduce an additional term to the objective, namely the closeness $\CostLength$ to a default configuration $\qdefault$:
\begin{align}
\CostLength(\q) = \frac{1}{2} \sum_{i=1}^{\NDoF}(\q_i - \qdefault_i)^2.
\label{eq:CostLength}
\end{align}
Minimizing $\CostLength$ makes the mapping unique and ensures that the solutions are close to the default configuration, making motion planning to this configuration faster and easier.

In summary, in the overall objective $\CostTotal$, one part is concerned with the frame at the end-effector $\CostFrame$, and one part accounts for the collisions and additional objectives $\CostRest$.
\begin{align}
\CostTotal &= \CostFrame + \CostRest \label{eq:CostTotal} \\
\text{with} \,\, \CostFrame &= \CostFramePos + \lambdaFrameRot \CostFrameRot , \, \CostRest = \lambdaWorldCollision \CostWorldCollision + \lambdaSelfCollision \CostSelfCollision + \lambdaLength \CostLength. \label{eq:CostRest}
\end{align}

Note that the weighting factors can be normalized independent of 
the robot and environment and are mainly to ensure higher importance of the collision terms over the secondary objectives like length. 
With this formulation, the optimal configuration $\q^*$ and solution to the IK problem is the one with the lowest objective
\begin{align}
\q^* = \argmin_{\q} \CostTotal(\q).
\end{align}

\subsection{Solver with Nullspace Projection}\label{sec:Solver-with-Nullspace-Projection}
While this formulation as an optimization problem is complete and \cref{eq:CostTotal} is used to train the unsupervised networks in \cref{sec:UnsupervisedLearning}, it is often not efficient to solve this complex cost function jointly.
To weaken the impact of competing terms in the objective function, we solve the IK problem in two steps.

First, we solve the pure IK with a projection step to ensure the constraints at the end-effector $\CostFramePos$ and $\CostFrameRot$ are satisfied.
This root search can be solved by iteratively applying the pseudo-inverse of the end-effector constraints:
\begin{align}
\Delta p = [\CostFramePos(\q), \CostFrameRot(\q)] \\
J = [\frac{\partial \CostFramePos(\q)}{\partial \q}, \frac{\partial \CostFrameRot(\q)}{\partial \q}] \\
\q_{i+1} = \q_i + J^{\dagger} \Delta p \label{eq:TCP_Projection}
\end{align}
For the humanoid Agile Justin, the IK requirements in the real world are to be accurate below \SI{e-4}{\meter} and  and \SI{e-3}{\radian}.
This numerical threshold is one order of magnitude more accurate than the actual accuracy of the calibrated system~\cite{Tenhumberg2021Elastic, Tenhumberg2022RGB}.

\begin{figure}[t]
    \centering
	\includegraphics[width=\linewidth]{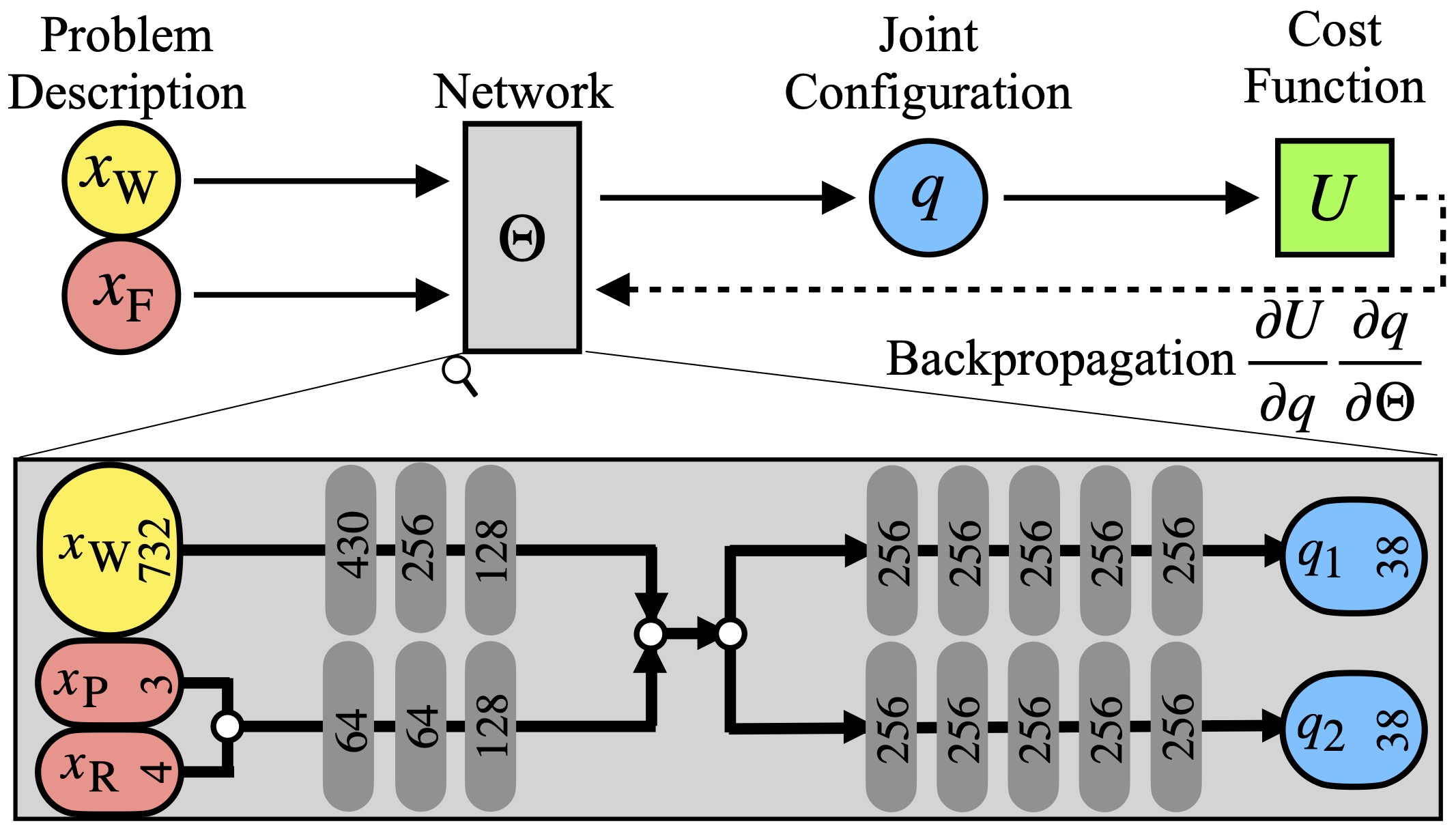}
	\caption{The graphic shows the flow of information through the neural network.
The IK problem is described by a world $x_{\text{w}}$, and a frame in the workspace $x_{\text{f}}$ and the network should predict a collision-free joint configuration that satisfies the end-effector. 
The dotted line indicates the backpropagation during unsupervised training, where the network weights $\NetWeights$ are directly updated according to the gradient of the cost function $\CostTotal$.
In the lower half, the detailed network structure for inverse kinematics of Agile Justin (19 Dof) is shown, with two heads and the unit vector representation for the joint angles described in \cref{sec:Learning-the-Inverse-Kinematics}.}
	\label{fig:information_flow_and_net}
\end{figure}

In the second step, we apply gradient descent with nullspace projection to satisfy the collision constraints and optimize the additional terms in $\CostRest$.
Each gradient step is again projected on the IK manifold to ensure the constraints at the end-effector stay satisfied
\begin{align}
\q_{i+1} = \q_i + (I - J^T  (J^T)^{\dagger}) \frac{\partial \CostRest(\q)}{\partial \q}.  \label{eq:Nullspace_GD}
\end{align}
These update steps push the configuration out of collision and closer to the default pose.
While this approach is straight forward to implement and converges quickly for a given sample, it is susceptible to the initial guess.
Especially for complex robots and environments, multiple samples are necessary until a feasible solution is found.

\section{Learning the Inverse Kinematics}\label{sec:Learning-the-Inverse-Kinematics}

The idea is to mitigate the strong dependence on the initial guess by using the prediction of a neural network as a warm-start for the optimization-based IK solver.
In this work, we compare two different learning approaches:
First, a supervised learning approach that relies on training data generated by the solver described in \cref{sec:Solver-with-Nullspace-Projection}.
Furthermore, we introduce an unsupervised regression approach, where the objective function given by \cref{eq:CostTotal} is directly used to update the network weights via backpropagation (see \cref{fig:information_flow_and_net}).
We use the same overall architecture for the supervised and unsupervised networks to compare the approaches.

\subsection{Environment Representation}\label{sec:WorldRepresentation}
\begin{figure}[t]
    \centering
    \includegraphics[width=0.95\linewidth]{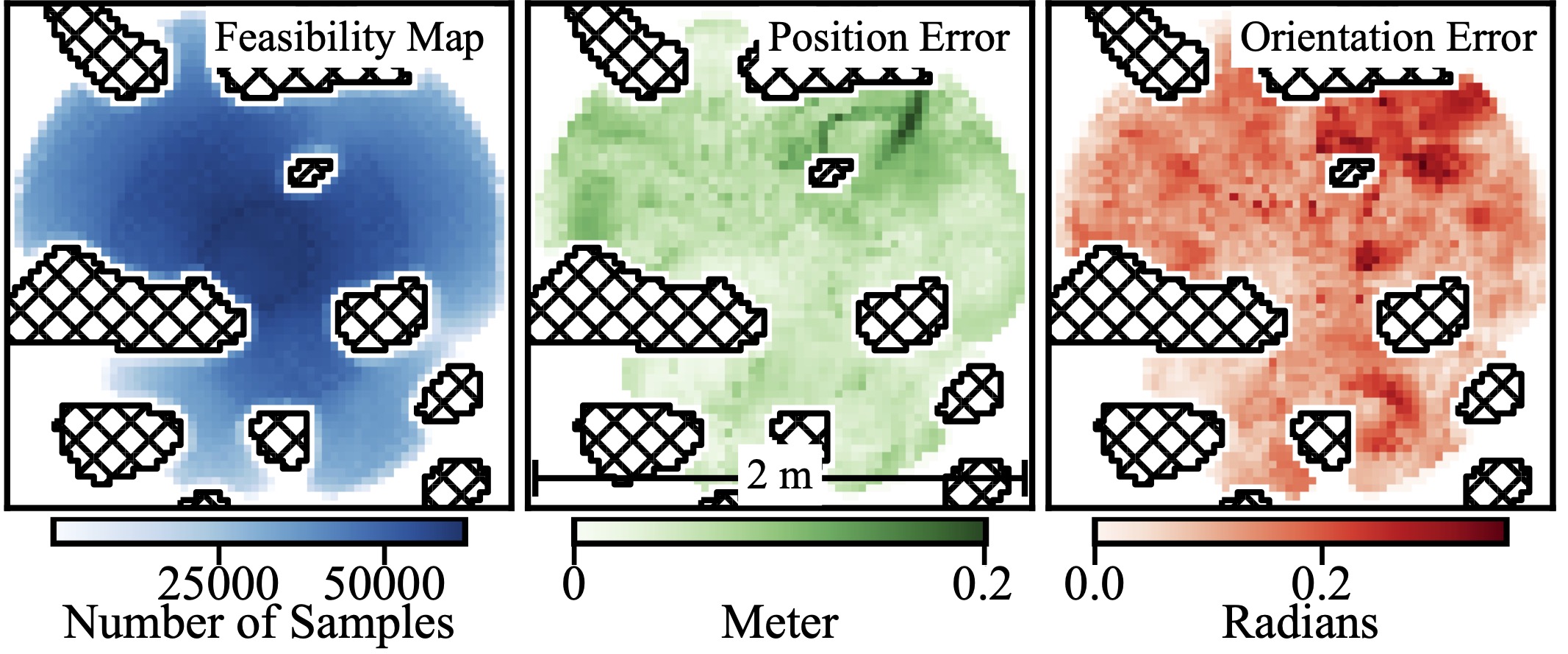}
    \caption{Feasibility map \textbf{(right/blue)} for the 2D arm with 5\,DoF for a specific environment. 
    Depending on the robot's kinematics, not only the parts of the workspace with obstacles are unreachable, but also areas behind obstacles.
    The overall number of feasible poses decreases towards the borders of the workspace.
    The maximal position error \textbf{(center/green)} and the maximal orientation error \textbf{(left/red)} highlight which regions are challenging for the network in more detail. 
    The  error maps show the maximal error over all orientations for each 2D position in the image.}
    \vspace{-0.2cm}
    \label{fig:StaticArm05_error-maps}
\end{figure}
As collision avoidance with the environment is a central aspect of the problem, the following section describes how to generate challenging training worlds and encode the scene to feed it into the networks.
The worlds were generated using Simplex noise \cite{Perlin2001}, as described by \citet{Tenhumberg2022} for motion planning. 
By adjusting the noise frequency and the threshold, we can create diverse and challenging environments for the robots. 
Examples of the different worlds can be seen in \cref{fig:ThreeRobots}.

To encode the environment for the networks, we use the BPS ~\cite{Prokudin2019}, which was already successfully used for robotic motion planning~\cite{Tenhumberg2022} between join configurations.
The advantage of this representation over occupancy grids and point clouds is that it is more memory-efficient, computationally efficient, and inherently permutation invariant.
The BPS representation can be understood as a subsampled SDF.
Formally, the BPS is an arbitrary but fixed set of points in the workspace $\BPS = \{\bps_i\}_{i=1}^{\Nbasisset}$.
The feature vector for the world $\xWorld$ passed to the network consists of the distances to the closest point in the environment for all basis points.
If a distance field $\distfield$ describes the environment, one can directly look up the feature vector
\begin{align}
\xWorld = [\distfield(\bps_i), \dots , \distfield(\bps_{\Nbasisset})].
\end{align}

\subsection{Supervised Learning}\label{sec:SupervisedLearning}
We use the algorithm described in \cref{sec:IK} for the sample generation.
Exhaustive multi-starts guarantee that a feasible solution is found, even in challenging scenes.
The supervised training relies on consistent data, implying the labels are all globally optimal. 
Ensuring this requires a lot of computational resources.
We use an efficient and generic cleaning method that uses the objective $\CostTotal$
and the current network to ensure all labels are close to the global optimum.
This cleaning is explained and analyzed for motion planning between joint configurations in \citet{Tenhumberg2022}.
After the data generation, we use a standard Mean Squared Error (MSE) loss to train the network supervised on the ground truth labels.

\subsection{Unsupervised Learning}\label{sec:UnsupervisedLearning}
Alternatively, as the objective function \cref{eq:CostTotal} holds all the necessary information to quantify a given configuration, it can be directly used as a loss function for training a network. 
\citet{Pandy2020} introduced unsupervised regression networks for robotic motion planning.
We adapt the idea and discuss the extensions needed in the context of IK.

For a given problem defined by a world $\xWorld$ and a frame $\xFrame$, one can directly calculate the gradients of \cref{eq:CostTotal} with respect to the network weights $\NetWeights$ by using the chain rule:
\begin{align}
\frac{\partial \CostTotal}{\partial \NetWeights} = \frac{\partial \CostTotal}{\partial \q} \frac{\partial \q}{\partial \NetWeights}.
\end{align}
In \cref{fig:information_flow_and_net}, the information flow through the network and the updates via backpropagation are shown.
The huge advantage of this unsupervised approach is that no computationally expensive generation of expert data is needed as in supervised learning. 
Here, different worlds $\xWorld$ and target frames $\xFrame$ are sampled randomly and via backpropagation the resulting gradients can be directly computed. 
This makes it faster and more straight forward to train.

\subsection{Learning and Network Architecture}\label{sec:Learning-the-Inverse-Kinematics-and-Network-Architecture}

To analyze the IK problem in the whole workspace, we generated feasibility and error maps of the robots in the different scenes.
\cref{fig:StaticArm05_error-maps} shows three maps:  feasibility (blue),  maximal position error (green), and maximal orientation error (red) for a given position in the workspace.
The maps were generated by sampling the whole joint space and collecting which euclidian targets were reached. 
Then, the position and the orientation error for each feasible target were computed.
Those maps can assess the network's performance over the whole workspace and are far more detailed than random test sets, which are commonly used.
The following sub-sections discuss the insights of this detailed analysis, which gave rise to our network and learning architecture.

\subsubsection{Boosting}\label{sec:Boosting}

\begin{figure}[t]
    \centering
    \includegraphics[width=\linewidth]{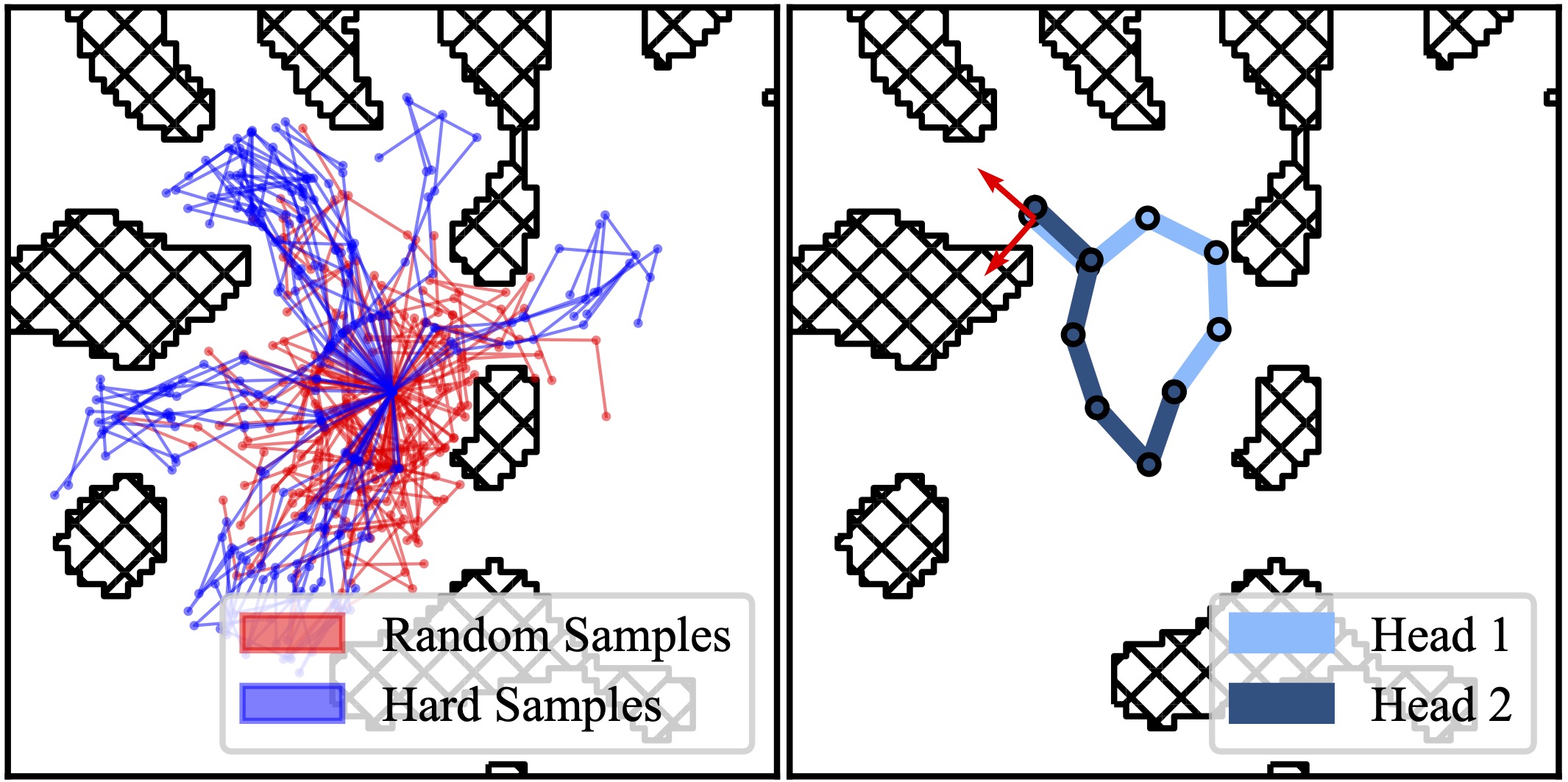}
    \caption{In the left image, 50 random but feasible samples for the robot in the given environment are drawn in red, and in blue, 50 samples that were in the hard set after the training finished (see \cref{sec:Boosting}).
    The challenging samples are more extended and fill the narrow passages in the world better than the random samples.
    In the right image, the predictions of the twin heads for a random sample are shown.
    While satisfying the end-effector, the two configurations show two distinct modes.}
    \label{fig:StaticArm05_hard-heads}
\end{figure}

\cref{fig:StaticArm05_error-maps} shows that the challenging samples close to obstacles are underrepresented if sampled randomly.
Random sampling tends to cluster in the central region and under-represent extreme positions which the robot can only reach fully extended.
We introduce a boosting technique to overcome this and produce reasonable initial guesses in challenging situations.
The idea is to have a set of challenging samples from which the training samples are chosen periodically. 
Similar to the method described by \citet{Tenhumberg2022}, we use the objective function $\CostTotal$ to over-represent complex samples.
We define a sample $q$ as hard if its cost \CostTotal(q) is four times higher than the rolling mean.

The effect of boosting can be seen in \cref{fig:StaticArm05_hard-heads}. 
Here, 50 samples are shown in blue, which were in the hard set after the training ended. 
In contrast, in red, 50 randomly sampled configurations are shown. 
Those are more clustered towards the center of the world.
This behavior can also be seen in \cref{fig:StaticArm05_error-maps}(left), where only a tiny fraction of the samples in the configuration space reach the borders of the workspace.

\subsubsection{Unit Vector Output}
We use a singularity-free representation for the networks' output using 2D unit vectors instead of the joint values in radians. 
In the plane, the unit vector is a natural representation of an angle, which inherently corresponds to the directions vector in the workspace.
This modification is especially relevant if the joint limits are $[-\pi, +\pi]$ or close to it.
However, also in 3D and with stricter joint limits, the network can easier represent the underlying problem when choosing this encoding.
The singularity-free representation omits the need for the network to internally represent a switch for joint values close to the singularity. 
We enforce this encoding by mapping the outputs of the network onto directly onto the unit circle.

\subsubsection{Twin-Headed Network}

\begin{figure}[t]
    \centering
    \includegraphics[width=\linewidth]{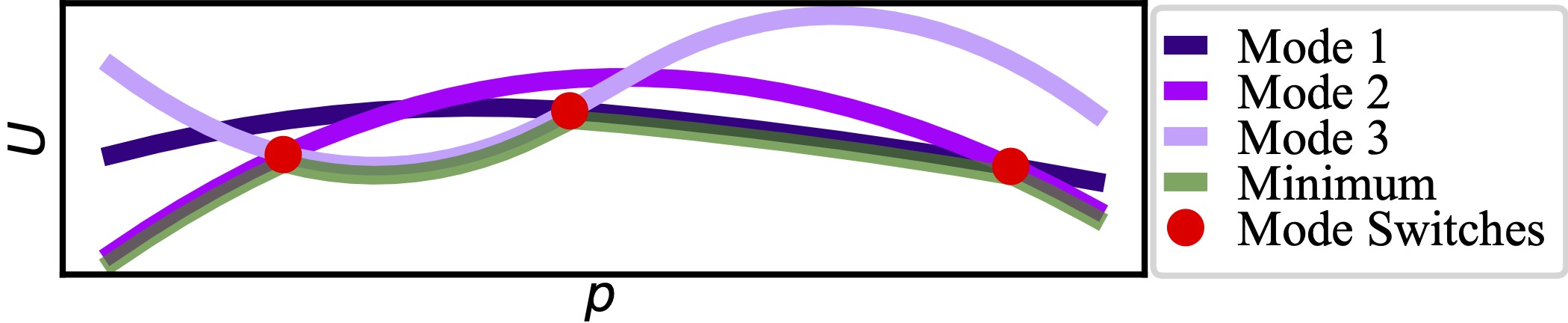}
    \caption{The 1D scheme of the optimization-based IK problem shows the necessity of mode switches over the workspace to get a globally optimal solution.
    Different modes exist with varying costs $\CostTotal$ over the workspace. 
    If the network should make the optimal prediction at each position $p$, it needs to switch between those modes.
    The transition regions are hard to represent for a neural network and can lead to significant errors (see \cref{fig:StaticArm05_why-twin}).}
    \label{fig:mode_switches_1d}
\end{figure}

\begin{figure}[t]
    \centering
    \includegraphics[width=0.95\linewidth]{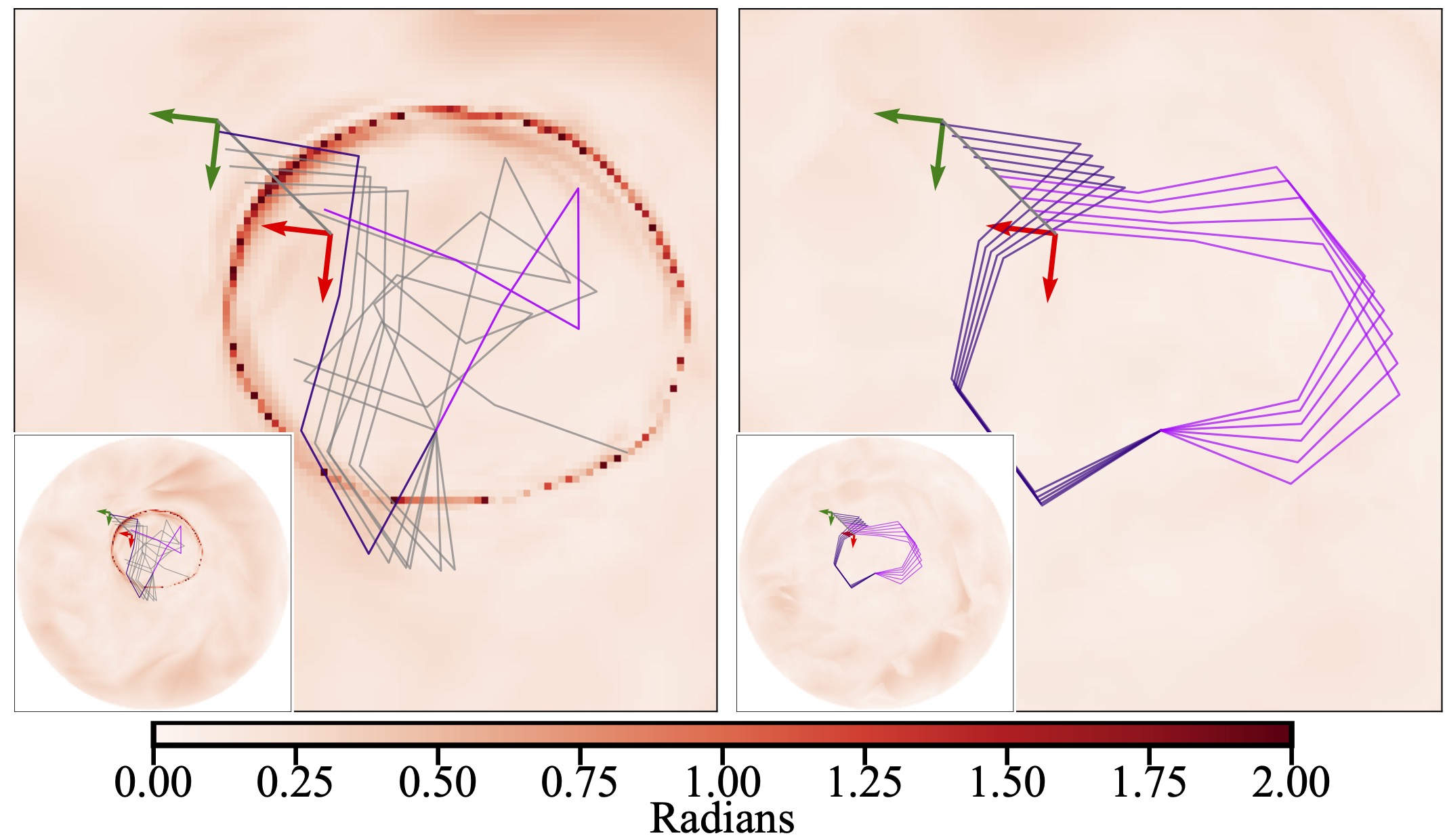}
    \caption{This comparison is between a single-headed network (left) and a twin-headed network (right) for the IK prediction of a 5\,DoF robot.
    The underlying red heat map indicates the worst orientation error across all $2\pi$ possible (discretized with 2880) goal orientations at each position.
    The distinct circular pattern (left) shows the transition region between two modes, where the prediction of the single-headed network breaks down.
    Moving the target frame at a specific orientation between those two regions leads to entirely wrong predictions.
    Each head of the twin model also has switching points, but as those two regions do not intersect, it can always predict valid and smoothly changing configurations (right).
    Visit the website for additional visualizations of the mode switches.}
    \label{fig:StaticArm05_why-twin}
\end{figure}

While the length cost $\CostLength$ ensures, in general, that there is one optimal solution, there are still different modes over the workspace, and the network must switch between those modes to successfully predict \textit{optimal} IK solutions for all possible targets.
\cref{fig:mode_switches_1d} visualizes the general concept of mode switches between a pair of modes to ensure an optimal solution. 

However, the network's prediction can become entirely wrong in these transition regions.
This behavior plus our solution is visualized in \cref{fig:StaticArm05_why-twin}.
For the 2D arm with 5\,DoF, the transition regions can be seen in the heat map of the maximal orientation error.
The specific position of these transition regions depends even on the initial weights of the network, but each initialization has the same behavior.
There are regions where the network needs to represent the switching between two modes. 
One can see the prediction breakdown by gradually moving the target frame from outside the ring (green) along a straight path to a position inside the ring (red).
In the transition region, the network switches modes and cannot produce valuable predictions.

By adding a second head to the network, which outputs a second prediction, one can overcome this problem.
Each head of the twin model also has its own transition regions, but as those two areas do not intersect, one always has a valid and smooth prediction for the configuration.
It is essential to add that two heads are enough, even for more complex settings with multiple modes.
The two heads do not represent the modes directly but only mask the transition region between pairs of modes. 

We introduce an additional loss $U_{\text{H}}=\|q_{\text{a}} - q_{\text{b}}\|$ between the two heads of the network to counteract mode collapse and gain a valuable second guess.
Both heads are trained simultaneously via back-propagation. 
\cref{fig:StaticArm05_hard-heads} (right) highlights that maximizing the difference in configuration space between those heads produces fundamentally different solution modes.  
Besides allowing sharp switches between modes, this approach leads to the simplest version of a generative model, with much more accessible training and no need for network ensembles~\cite{Lembono2021} to prevent mode collapse.

\section{Results}\label{sec:results}

\begin{figure}[t]
    \centering
    \includegraphics[width=\linewidth]{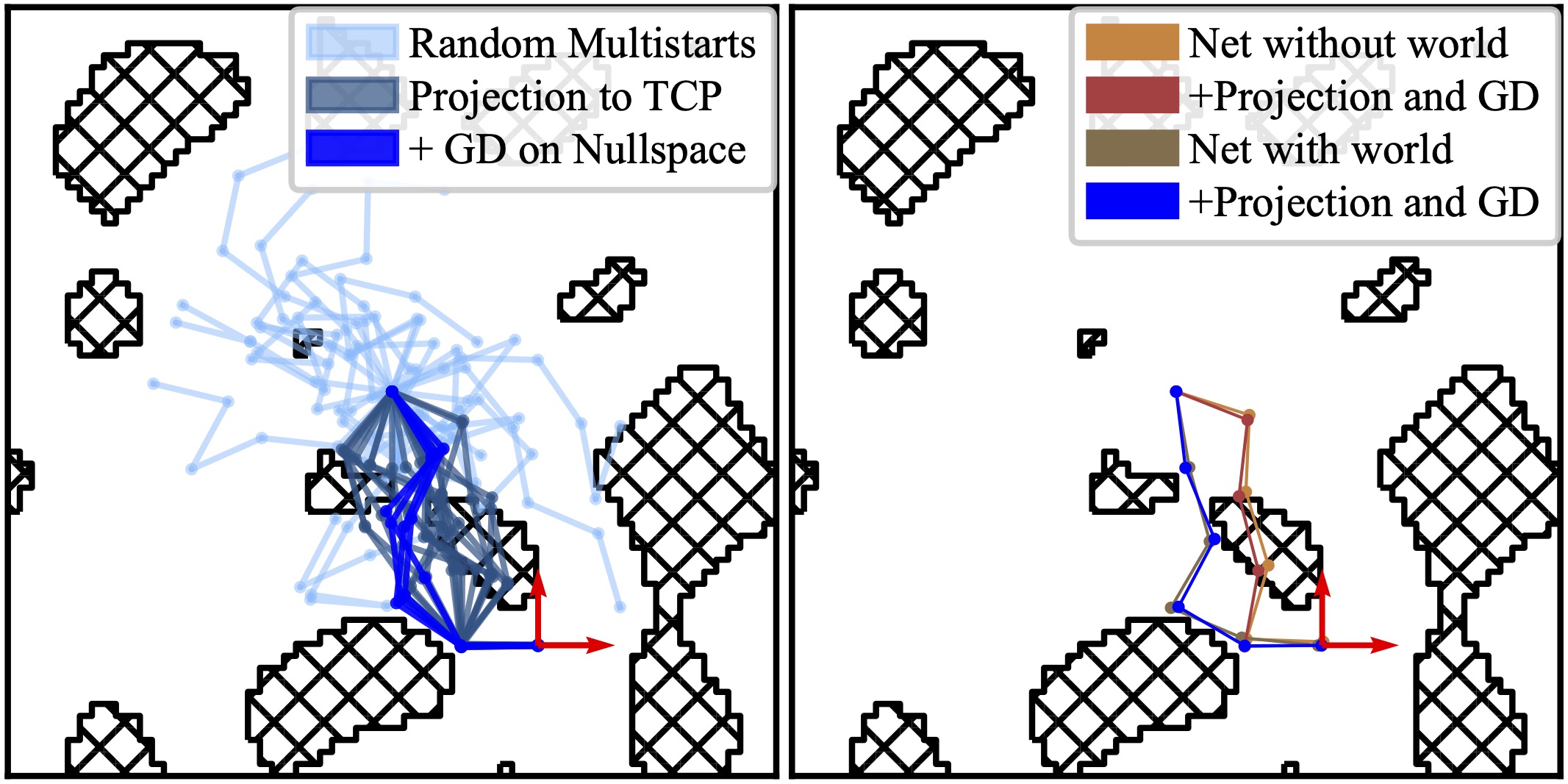}
    \caption{The graph visualizes the collision-free IK solution process.
    In the \textbf{left image}, 20 initial random guesses of configurations (see legend for used colors) are used. 
    These configurations are then projected onto the desired TCP (red coordinate system) using \cref{eq:TCP_Projection}, ensuring that translational and rotational constraints are satisfied. 
    Then, we compute the gradient of $\CostRest$ and apply Gradient Descent inside the TCP-nullspace \cref{eq:Nullspace_GD} to move the robot out of collision and closer to the default configuration.
    After these steps, only five feasible solutions remain.
    The \textbf{right image} showcases the same two-stage process when using the (single)  prediction of an IK network but for two network variants. One network is trained without the world as an input, while the other network incorporates the BPS of the world to predict collision-free IK solutions. 
    As the predictions of both networks are close to the desired TCP, the pure projection step on the TCP is not shown here. 
    However, only the prediction of the world-aware network converges to a feasible solution while the other gets stuck in a collision.}
    \label{fig:StaticArm05_random-nets}
\end{figure}

First, we demonstrate the effectiveness of our approach for the problem of collision-free IK in the case of a 2D robotic arm. \cref{fig:StaticArm05_random-nets} shows the steps of our IK procedure and compares it to using simple random sampling for generating initial guesses. 
Only five of the initial 20 configurations are feasible after both optimization steps.
This ratio gets even worse for more complex robots in challenging 3D environments, which \cref{tab:results} analyzes in more detail.

The right-hand figure shows the same two-stage procedure for two network predictions.
One trained without the world as a dedicated input and one which uses the BPS of the world to predict collision-free IK solutions.
One can see clearly how close the two predictions are to the desired TCP. 
Furthermore, the prediction of the world-aware network is already in the correct narrow passage between the obstacles. 
Using this prediction as an initial guess eliminates the need for multi-starts in most cases and leads to quicker convergence, as the optimizer only needs a few iterations for a feasible solution. 

\subsection{Experiments}

This section shows the results for the supervised and unsupervised learning methods for multiple robots with different complexity.
All timings are measured on a computer with Intel i9-9820X @ 3.30\,GHz with 32\,GB RAM. 
While all 16 cores are used for training, the online prediction runs only on a single core.
To evaluate the networks, we use their prediction as a warm-start for the optimization-based solver described in \cref{sec:Solver-with-Nullspace-Projection} and compare convergence and feasibility rates for unseen test sets.

\cref{tab:ablation_study} shows an ablation study for the learning and network architecture proposed in \cref{sec:Learning-the-Inverse-Kinematics-and-Network-Architecture}.
For the humanoid robot Agile Justin, the different networks were trained on 300 random worlds and evaluated on 20 unseen worlds drawn from the same distribution. 
The size of the test set was 100000 samples. 
Because not the network prediction directly is used on the robot but the converged result, we report the feasibility rate after 10 iterations of the solver.
The table shows that each architectural component improves the performance of the network. 
In the extreme case where none of those methods are used, the feasibility rate is only 25\%, while the final performance is close to 100\%.
Notably, the boosting does not improve the mean performance but significantly reduces the maximal error of the network's predictions.
As this approach over-represents the complex samples with a large objective $\CostTotal$, it is designed to improve those worst cases.
This design is crucial if one uses those network predictions as a warm-start for an optimization-based solver in challenging scenes: 
Long searches with many multi-starts slow down the numerical solver for those cases.

\begin{table}[t]
    \caption{Training times for the different Networks}
    \label{tab:training_times}
    \centering
    \begin{tabular}{c|cc|c}
\toprule
            & \multicolumn{2}{c|}{Supervised}    & Unsupervised \\
Robots      & Data Generation   & Training       & Training     \\
\midrule
\StaticArm  & 34.6\,h           & 2.1\,h          & 2.6\,h      \\
\JustinArm  & 71.3\,h           & 2.7\,h          & 3.0\,h      \\
\Justin     & 95.4\,h           & 5.4\,h          & 6.9\,h      \\
\bottomrule 
\end{tabular}

\end{table}

\begin{table}[t]    
    \caption{Ablation Study of the Network Prediction for Agile Justin}
    \label{tab:ablation_study}
    \centering
    \begin{tabular}{ccc|c}
\toprule
Training w. & Twin-Headed & Unit Vector & Feasibility \\  
Boosting    & Network     & Output      &             \\  \hline 
Yes         & Yes         & Yes         & 0.986       \\  
Yes         & Yes         & No          & 0.871       \\  
Yes         & No          & Yes         & 0.695       \\  
Yes         & No          & No          & 0.596       \\  
No          & Yes         & Yes         & 0.781       \\  
No          & Yes         & No          & 0.741       \\  
No          & No          & Yes         & 0.569       \\  
No          & No          & No          & 0.248       \\  
\bottomrule
\end{tabular}

\end{table}

\begin{table*}[t]
    \caption{Feasibility and Convergence for the different sampling modes for the warm-start of the IK Solver}
    \label{tab:results}
    \centering
    \begin{tabular}{ccc|cccc}
\toprule
Robots                      & DoF  & Initial Guess & Avg. Mulit-Starts {[}\#{]}  & Feasibility (1) {[}\%{]} & Avg. Iterations {[}\#{]} & Avg. Length Cost $\CostLength$ [rad] \\
\midrule
\StaticArm                  &      & Random        & $13.27 \pm 5.41$            & 19.7                     & $12.87 \pm 3.78$         & $4.19 \pm 0.75$       \\
Random World                & 5    & Supervised    & $3.64  \pm 1.29$            & 81.3                     & $9.93  \pm 3.67$         & $3.48 \pm 0.71$       \\
                            &      & Unsupervised  & $3.35  \pm 1.37$            & 83.4                     & $8.53  \pm 3.29$         & $3.41 \pm 0.69$       \\
\midrule
\JustinArm                  &      & Random        & $17.57 \pm 4.53$            & 14.4                     & $15.69 \pm 2.89$         & $3.36 \pm 0.68$       \\
Shelf World                 & 7    & Supervised    & $2.97  \pm 0.61$            & 88.7                     & $9.31  \pm 3.78$         & $2.91 \pm 0.70$       \\
                            &      & Unsupervised  & $3.06  \pm 0.55$            & 92.6                     & $8.76  \pm 4.01$         & $2.85 \pm 0.65$       \\
\midrule
\Justin                     &      & Random        & $24.52 \pm 7.18$            & 8.3                      & $13.88 \pm 3.93$         & $6.56 \pm 0.93$       \\
Shelf World                 & 19   & Supervised    & $4.41  \pm 0.94$            & 88.9                     & $6.91  \pm 3.66$         & $4.72 \pm 0.41$       \\
                            &      & Unsupervised  & $4.29  \pm 0.97$            & 87.6                     & $7.25  \pm 3.61$         & $4.10 \pm 0.53$       \\
\bottomrule
\end{tabular}
\end{table*}

The results of comparing the supervised and unsupervised network against a randomly sampled initial guess are summarized in \cref{tab:results}.
This evaluation was performed for three robots: A 2D Arm with 5\,DoF, the LWR III with 7\,DoF, and Agile Justin with 19\,DoF (see \cref{fig:ThreeRobots}).
In 3D, we used a shelf environment like depicted in \cref{fig:Justin19_shelf}.
Here 10000 target frames were randomly sampled in the respective boxes in the shelf. 
The overall orientation of the target frame was aligned with the shelf, and noise was added to ensure feasible yet challenging samples.
The shelf environment is closer to a real-world setting and has notably different attributes than the random worlds the networks were trained on. 

\cref{tab:results} shows that the average feasibility rate of the initial guesses from the networks outperforms the random baseline significantly for a single initial guess (denoted as (1)).
Furthermore, the average number of iterations to converge is also decreased. 
The overall speed advantage can be seen directly from the necessary iterations difference.
For the humanoid robot Agile Justin (19 DoF), the computation time for a single iteration is 0.8\,ms on our testing machine. 
This leads to an overall solve time of under 10ms for the collision-free IK in unseen environments.
The learned warm-starts outperform the random multi-starts in solving time, and the length cost \cref{eq:CostLength} is reduced.
These solutions are often more convenient and easier to integrate into larger motion planning tasks than random solutions. 

Besides the improvement of the learning-based approaches over the random multi-start, it can also be seen that supervised and unsupervised training perform similarly well.
Overall, this gives an advantage to the unsupervised method, as it requires far less time to train as no prior data generation and data cleaning~\cite{Tenhumberg2022} is needed as \cref{tab:training_times} shows. 

\subsection{Real-World Experiment on the Humanoid Agile Justin}
\begin{figure}[t]
    \centering
    \includegraphics[width=\linewidth]{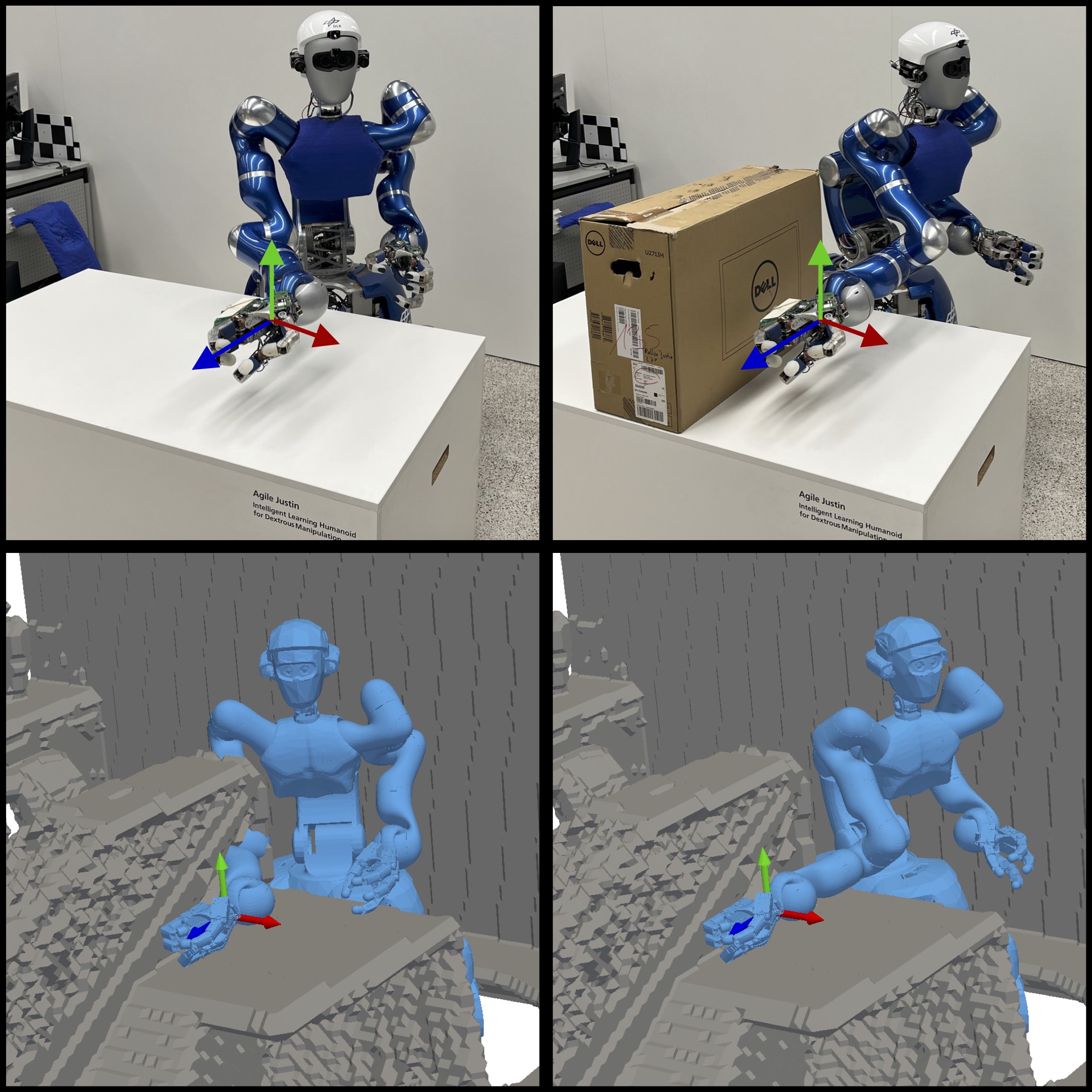}
    \caption{Difference between standard IK (left) and collision-free IK (right) for the humanoid robot Agile Justin in a real table scene. 
    The rendered images show the robots' self-acquired high-resolution voxel model ~\cite{Wagner2013} of the scene. 
    This conservative occupancy map was encoded with BPS and used as input for the neural network. 
    While only trained on random worlds, its prediction for this unseen world converges to a collision-free solution.}
    \label{fig:real_table_ik}
\end{figure}

We present real-world results on the humanoid Agile Justin to show the need for collision-free IK. 
\cref{fig:real_table_ik} shows two table scenes; the robot should move the right TCP to the same position in both cases, first without obstacle and then with an additional obstruction. 
The rendered images in the bottom row show the self-acquired high-resolution voxel model ~\cite{Wagner2013}. 
The optimal solution to the IK for the simple scene does collide with the additional obstacle. 
The whole arm is stuck in the box on the table, and using this solution as a warm-start for our solver does not converge to a collision-free solution.
However, using the neural network's prediction as an initial guess produces the solution shown on the right.
The BPS representation and the proposed training scheme were robust enough to generalize to high-resolution voxel models collected by Agile Justin's depth camera~\cite{Wagner2013}, even if the training was only on random simplex worlds.

\section{Conclusions and Future Work}

We introduced an unsupervised training method for learning the IK with collision avoidance. 
It works for a humanoid robot with 19\,DoF in challenging and diverse environments sensed with its integrated 3D sensor. 
An IK solution with an accuracy of \SI{e-4}{\meter} and \SI{e-3}{\radian} is computed in only \SI{10}{\milli\second} on a single CPU core.
Our method trains ten times faster than supervised training by avoiding the generation of an exhaustive training data set. 
It massively outperforms a multi-start baseline, as we showed in an elaborate benchmark with multiple robots in challenging environments.
Based on a detailed analysis of the IK problem with collision avoidance, we derived our network and learning architecture with boosting to enable rare-case performance and dual-heads to handle the necessary switching between different configuration modes. 
An ablation study demonstrates the relevance of this architecture.

Separating the task of grasping a specific object in a given scene into the subtasks of finding a stable grasp, getting the end configuration via IK, which allows this grasp, and then planning from a start point to that configuration is not always possible. 
Future work will integrate the IK tighter into the related grasping and path-planning problems. 
Ideally, grasping a specific object in a given scene must be solved jointly, as this guarantees the feasibility of the complete task and allows us to find globally optimal solutions.

\footnotesize
\bibliographystyle{IEEEtranN-modified}
\bibliography{IEEEabrv, references2.bib}
\end{document}